\def\year{2018}\relax
\documentclass[letterpaper]{article} 
\usepackage{aaai18} 
\usepackage{times} 
\usepackage{courier} 
\usepackage{url} 
\usepackage[scaled=.86]{helvet}

\usepackage[colorlinks=true,citecolor=blue,urlcolor=black,linkcolor=red]{hyperref}
\usepackage[pdftex]{graphicx}
\usepackage{subfig}
\usepackage{algorithm}
\usepackage[noend]{algpseudocode}

\algnewcommand{\IfInline}[2]{\State \algorithmicif\ #1 \algorithmicthen\ ~#2~}
\usepackage{amsmath}
\usepackage{amssymb}
\usepackage{amsthm}

\DeclareMathOperator{\MLP}{MLP}
\DeclareMathOperator{\RNN}{RNN}
\newcommand*\diff{\mathop{}\!\mathrm{d}}

\newcommand*\ps[1]{\mathrm{#1}} 
\usepackage{bm}
\usepackage[noamssymbols,slantedGreek]{newtxmath}
\usepackage[cal=euler,calscaled=1.0,scr=boondoxo,scrscaled=1.0,bb=px]{mathalfa}
\usepackage{makecell}
\newcommand\Tstrut{\rule{0pt}{2.6ex}}         
\newcommand\Bstrut{\rule[-0.9ex]{0pt}{0pt}}   
\usepackage{xcolor}

\begin{document}
%
\title{A Neural Stochastic Volatility Model}
\author{Rui Luo\textsuperscript{\dag}, Weinan Zhang\textsuperscript{\ddag}, Xiaojun Xu\textsuperscript{\ddag}, and Jun Wang\textsuperscript{\dag}\\
\textsuperscript{\dag}University College London and \textsuperscript{\ddag}Shanghai Jiao Tong University\\
\texttt{\{r.luo,j.wang\}@cs.ucl.ac.uk,~\{wnzhang,xuxj\}@apex.sjtu.edu.cn}
}
\maketitle
\begin{abstract}
In this paper, we show that the recent integration of statistical models with deep recurrent neural networks provides a new way of formulating volatility (the degree of variation of time series) models that have been widely used in time series analysis and prediction in finance. The model comprises a pair of complementary stochastic recurrent neural networks: the generative network models the joint distribution of the stochastic volatility process; the inference network approximates the conditional distribution of the latent variables given the observables. Our focus here is on the formulation of temporal dynamics of volatility over time under a stochastic recurrent neural network framework.
Experiments on real-world stock price datasets demonstrate that the proposed model generates a better volatility estimation and prediction that outperforms mainstream methods, e.g., deterministic models such as GARCH and its variants, and stochastic models namely the MCMC-based model \emph{stochvol} as well as the Gaussian process volatility model \emph{GPVol}, on average negative log-likelihood.
\end{abstract}


\section{Introduction}

The volatility of the price movements reflects the ubiquitous uncertainty within financial markets. It is critical that the level of risk (aka, the degree of variation), indicated by volatility, is taken into consideration before investment decisions are made and portfolio are optimised \cite{hull2006options}; volatility is substantially a key variable in the pricing of derivative securities. Hence, estimating and forecasting volatility is of great importance in branches of financial studies, including investment, risk management, security valuation and monetary policy making \cite{poon2003forecasting}.

Volatility is measured typically by employing the standard deviation of price change in a fixed time interval, such as a day, a month or a year. The higher the volatility is, the riskier the asset should be. One of the primary challenges in designing volatility models is to identify the existence of latent stochastic processes and to characterise the underlying dependences or interactions between variables within a certain time span. A classic approach has been to handcraft the characteristic features of volatility models by imposing assumptions and constraints, given prior knowledge and observations. Notable examples include autoregressive conditional heteroscedasticity (ARCH) model \cite{engle1982autoregressive} and the extension, generalised ARCH (GARCH) \cite{bollerslev1986generalized}, which makes use of autoregression to capture the properties of time-varying volatility within many time series. As an alternative to the GARCH model family, the class of stochastic volatility (SV) models specify the variance to follow some latent stochastic process \cite{hull1987pricing}.
Heston \cite{heston1993closed} proposed a continuous-time model with the volatility following an Ornstein-Uhlenbeck process and derived a closed-form solution for options pricing. 
Since the temporal discretisation of continuous-time dynamics sometimes leads to a deviation from the original trajectory of system, those continuous-time models are seldom applied in forecasting. For practical purposes of forecasting, the canonical model \cite{jacquier2002bayesian,kim1998stochastic} formulated in a discrete-time fashion for regularly spaced data such as daily prices of stocks is of great interest.
While theoretically sound, those approaches require strong assumptions which might involve detailed insight of the target sequences and are difficult to determine without a thorough inspection.

In this paper, we take a fully data driven approach and determine the configurations with as few exogenous input as possible, or even purely from the historical data. We propose a neural network re-formulation of stochastic volatility by leveraging stochastic models and recurrent neural networks (RNNs).
In inspired by the work from Chung et al. \cite{DBLP:conf/nips/ChungKDGCB15} and Fraccaro et al. \cite{DBLP:conf/nips/FraccaroSPW16}, the proposed model is rooted in variational inference and equipped with the latest advances of stochastic neural networks. The model inherits the fundamentals of SV model and provides a general framework for volatility modelling; it extends previous sequential frameworks with autoregressive and bidirectional architecture and provide with a more systematic and volatility-specific formulation on stochastic volatility modelling for financial time series.
We presume that the latent variables follow a Gaussian autoregressive process, which is then utilised to model the variance process. Our neural network formulation is essentially a general framework for volatility modelling, which covers two major classes of volatility models in financial study as the special cases with specific weights and activations on neurons.

Experiments with real-world stock price datasets are performed. The result shows that the proposed model produces more accurate estimation and prediction, outperforming various widely-used deterministic models in the GARCH family and several recently proposed stochastic models on average negative log-likelihood; the high flexibility and rich expressive power are validated.


\section{Related Work}

A notable framework for volatility is autoregressive conditional heteroscedasticity (ARCH) model \cite{engle1982autoregressive}: it can accurately identify the characteristics of time-varying volatility within many types of time series. Inspired by ARCH model, a large body of diverse work based on stochastic process for volatility modelling has emerged \cite{bollerslev1994arch}. Bollerslev \cite{bollerslev1986generalized} generalised ARCH model to the generalised autoregressive conditional heteroscedasticity (GARCH) model in a manner analogous to the extension from autoregressive (AR) model to autoregressive moving average (ARMA) model by introducing the past conditional variances in the current conditional variance estimation. Engle and Kroner \cite{engle1995multivariate} presented theoretical results on the formulation and estimation of multivariate GARCH model within simultaneous equations systems. The extension to multivariate model allows the covariance to present and depend on the historical information, which are particularly useful in multivariate financial models. An alternative to the conditionally deterministic GARCH model family is the class of stochastic volatility (SV) models, which first appeared in the theoretical finance literature on option pricing \cite{hull1987pricing}. The SV models specify the variance to follow some latent stochastic process such that the current volatility is no longer a deterministic function even if the historical information is provided. As an example, Heston's model \cite{heston1993closed} characterises the variance process as a Cox-Ingersoll-Ross process driven by a latent Wiener process. While theoretically sound, those approaches require strong assumptions which might involve complex probability distributions and non-linear dynamics that drive the process.
Nevertheless, empirical evidences have confirmed that volatility models provide accurate prediction \cite{andersen1998answering} and models such as ARCH and its descendants/variants have become indispensable tools in asset pricing and risk evaluation. 
Notably, several models have been recently proposed for practical forecasting tasks: Kastner et al. \cite{DBLP:journals/csda/KastnerF14} implemented the MCMC-based framework \emph{stochvol} where the ancillarity-sufficiency interweaving strategy (ASIS) is applied for boosting MCMC estimation of stochastic volatility; Wu et al. \cite{DBLP:conf/nips/WuHG14} designed the \emph{GP-Vol}, a non-parametric model which utilises Gaussian processes to characterise the dynamics and jointly learns the process and hidden states via online inference algorithm. Despite the fact that it provides us with a practical approach towards stochastic volatility forecasting, both models require a relatively large volume of samples to ensure the accuracy, which involves very expensive sampling routine at each time step. Another drawback is that those models are incapable to handle the forecasting task for multivariate time series. 

On the other hand, deep learning \cite{DBLP:journals/nature/LeCunBH15,DBLP:journals/nn/Schmidhuber15} that utilises nonlinear structures known as deep neural networks, powers various applications. It has triumph over pattern recognition challenges, such as image recognition \cite{DBLP:conf/nips/KrizhevskySH12}, speech recognition \cite{DBLP:conf/nips/ChorowskiBSCB15}, machine translation \cite{DBLP:journals/corr/BahdanauCB14} to name a few.

Time-dependent neural networks models include RNNs with neuron structures such as long short-term memory (LSTM) \cite{DBLP:journals/neco/HochreiterS97}, bidirectional RNN (BRNN) \cite{DBLP:journals/tsp/SchusterP97}, gated recurrent unit (GRU) \cite{DBLP:conf/emnlp/ChoMGBBSB14} and attention mechanism \cite{DBLP:conf/icml/XuBKCCSZB15}.
Recent results show that RNNs excel for sequence modelling and generation in various applications \cite{DBLP:conf/icml/OordKK16,DBLP:conf/emnlp/ChoMGBBSB14,DBLP:conf/icml/XuBKCCSZB15}. However, despite its capability as non-linear universal approximator, one of the drawbacks of neural networks is its deterministic nature. Adding latent variables and their processes into neural networks would easily make the posteriori computationally intractable. Recent work shows that efficient inference can be found by variational inference when hidden continuous variables are embedded into the neural networks structure \cite{DBLP:journals/corr/KingmaW13,DBLP:conf/icml/RezendeMW14}. Some early work has started to explore the use of variational inference to make RNNs stochastic:
Chung et al. \cite{DBLP:conf/nips/ChungKDGCB15} defined a sequential framework with complex interacting dynamics of coupling observable and latent variables whereas Fraccaro et al. \cite{DBLP:conf/nips/FraccaroSPW16} utilised heterogeneous backward propagating layers in inference network according to its Markovian properties.

In this paper, we apply the stochastic neural networks to solve the volatility modelling problem. In other words, we model the dynamics and stochastic nature of the degree of variation, not only the mean itself. Our neural network treatment of volatility modelling is a general one and existing volatility models (e.g., the Heston and GARCH models) are special cases in our formulation.


\section{Preliminaries: Volatility Models}

Volatility models characterise the dynamics of volatility processes, and help estimate and forecast the fluctuation within time series. As it is often the case that one seeks for prediction on quantity of interest with a collection of historical information at hand, we presume the conditional variance to have dependency -- either deterministic or stochastic -- on history, which results in two categories of volatility models.


\subsection{Deterministic Volatility Models: the GARCH Model Family}

The GARCH model family comprises various linear models that formulate the conditional variance at present as a linear function of observations and variances from the past. Bollerslev's extension \cite{bollerslev1986generalized} of Engle's primitive ARCH model \cite{engle1982autoregressive}, referred as generalised ARCH (GARCH) model, is one of the most well-studied and widely-used volatility models:
\begin{align}
\label{eq:garch}
\sigma^2_t &= \alpha_0 + \sum_{i=1}^p \alpha_i x^2_{t-i} + \sum_{j=1}^q \beta_j \sigma^2_{t-j}, \\
\label{eq:assumption}
x_t &\sim \mathscr{N}(0, \sigma^2_t),
\end{align}
where Eq. \eqref{eq:assumption} represents the assumption that the observation $x_t$ follows from the Gaussian distribution with mean 0 and variance $\sigma^2_t$; the (conditional) variance $\sigma^2_t$ is fully determined by a linear function (Eq. \eqref{eq:garch}) of previous observations $\{x_{<t}\}$ and variances $\{\sigma^2_{<t}\}$. Note that if $q=0$ in Eq. \eqref{eq:garch}, GARCH model degenerates to ARCH model.

Various variants have been proposed ever since. Glosten, Jagannathan and Runkle \cite{glosten1993relation} extended GARCH model with additional terms to account for asymmetries in the volatility and proposed GJR-GARCH model; Zakoian \cite{zakoian1994threshold} replaced the quadratic operators with absolute values, leading to threshold ARCH/GARCH (TARCH) models. The general functional form is formulated as
\begin{align}
\label{eq:tarch}
\sigma^d_t = \alpha_0 &+ \sum_{i=1}^p \alpha_i |x_{t-i}|^d + \sum_{j=1}^q \beta_j \sigma^d_{t-j}\notag \\
&+ \sum_{k=1}^o \gamma_k |x_{t-k}|^d I[x_{t-k} < 0],
\end{align}
where $I[x_{t-k} < 0]$ is the indicator function: $I[x_{t-k} < 0] = 1$ if $x_{t-k} < 0$, and 0 otherwise, which allows for asymmetric reactions of volatility in terms of the sign of previous $\{x_{<t}\}$. 

Various variants of GARCH model can be expressed by assigning values to parameters $p,o,q,d$ in Eq. \eqref{eq:tarch}:
{
\begin{enumerate}
\setlength\itemsep{0em}
\item{ARCH($p$): $p \in \mathbb{N}^+$; $q \equiv 0$; $o \equiv 0$; $d \equiv 2$}
\item{GARCH($p,q$): $p \in \mathbb{N}^+$; $q \equiv 0$; $o \equiv 0$; $d \equiv 2$}
\item{GJR-GARCH($p,o,q$): $p \in \mathbb{N}^+$; $q \in \mathbb{N}^+$; $o \in \mathbb{N}^+$; $d \equiv 2$}
\item{AVARCH($p$): $p \in \mathbb{N}^+$; $q \equiv 0$; $o \equiv 0$; $d \equiv 2$}
\item{AVGARCH($p,q$): $p \in \mathbb{N}^+$; $q \in \mathbb{N}^+$; $o \equiv 0$; $d \equiv 2$}
\item{TARCH($p,o,q$): $p \in \mathbb{N}^+$; $q \in \mathbb{N}^+$; $o \in \mathbb{N}^+$; $d \equiv 1$}
\end{enumerate}
}%

Another fruitful specification shall be Nelson's exponential GARCH (EGARCH) model \cite{nelson1991conditional}, which instead formulates the dependencies in log-variance $\log(\sigma^2_t)$:
\begin{align}
\label{eq:egarch}
\log(\sigma^2_t) &= \alpha_0 + \sum_{i=1}^p \alpha_i g(x_{t-i}) + \sum_{j=1}^q \beta_j \log(\sigma^2_{t-j}), \\
\label{eq:g}
g(x_t) &= \theta x_t + \gamma (|x_t| - \mathbb{E}[|x_t|]),
\end{align}
where $g(x_t)$ (Eq. \eqref{eq:g}) accommodates the asymmetric relation between observations and volatility changes. If setting $q \equiv 0$ in Eq. \eqref{eq:egarch}, EGARCH($p,q$) then degenerates to EARCH($p$).


\subsection{Stochastic Volatility Models}

An alternative to the (conditionally) deterministic volatility models is the class of stochastic volatility (SV) models. First introduced in the theoretical finance literature, earliest SV models such as Hull and White's \cite{hull1987pricing} as well as Heston model \cite{heston1993closed} are formulated by stochastic differential equations in a continuous-time fashion for analysis convenience. In particular, Heston model instantiates a continuous-time stochastic volatility model for univariate processes:
\begin{align}
\label{eq:heston_z}
\diff{\sigma}(t) &= -\beta \sigma(t) \diff{t} + \delta \diff{w^{\sigma}}(t), \\
\diff{x}(t) &= (\mu - 0.5\sigma^2(t)) \diff{t} + \sigma(t) \diff{w^{x}}(t).
\end{align}
where $x(t)=\log{s}(t)$ is the logarithm of stock price $s_t$ at time $t$, $w^{x}(t)$ and $w^{\sigma}(t)$ represent two correlated Wiener processes and the correlation between $\diff{w^{x}}(t)$ and $\diff{w^{\sigma}}(t)$ is expressed as $\mathbb{E}[\diff{w^{x}}(t) \cdot \diff{w^{\sigma}}(t)] = \rho \diff{t}$.

For practical use, empirical versions of the SV model, typically formulated in a discrete-time fashion, are of great interest. The canonical model \cite{jacquier2002bayesian,kim1998stochastic} for regularly spaced data is formulated as
\begin{align}
\label{eq:sv1}
\log(\sigma^2_t) &= \eta + \phi (\log(\sigma^2_{t-1})-\eta) + z_t, \\
\label{eq:sv2}
z_t &\sim \mathscr{N}(0, \sigma^2_{z}), \quad x_t \sim \mathscr{N}(0, \sigma^2_t).
\end{align}
Equation~\eqref{eq:sv1} indicates that the (conditional) log-variance $\log(\sigma^2_t)$ depends on not only the historical log-variances $\{\log(\sigma^2_t)\}$ but a latent stochastic process $\{z_t\}$. The latent process $\{z_t\}$ is, according to Eq. \eqref{eq:sv2}, white noise process with i.i.d. Gaussian variables.

Notably, the volatility $\sigma^2_t$ is no longer conditionally deterministic (i.e. deterministic given the complete history $\{\sigma^2_{<t}\}$) but to some extent stochastic in the setting of SV models: Heston model involves two correlated continuous-time Wiener processes while the canonical model is driven by a discrete-time Gaussian white-noise process.


\subsection{Volatility Models in a General Form}

Hereafter we denote the sequence of observations as $\{x_t\}$ and the latent stochastic process as $\{z_t\}$. As seen in previous sections, the dynamics of the volatility process $\{\sigma^2_t\}$ can be abstracted in the form of
\begin{align}
\label{eq:f}
\sigma^2_t = f(\sigma^2_{<t}, x_{<t}, z_{\le t}) = \Sigma^x(x_{<t}, z_{\le t}).
\end{align}

The latter equality holds when we recurrently substitute $\sigma^2_\tau$ with $f(\sigma^2_{<\tau}, x_{<\tau}, z_{\le \tau})$ for all $\tau<t$. For models within the GARCH family, we discard $z_{\le t}$ in $\Sigma^x(x_{<t}, z_{\le t})$ (Eq. \eqref{eq:f}); on the other hand, for the primitive SV model, $x_{<t}$ is ignored instead. We can loosen the constraint that $x_t$ is zero-mean to a time-varying mean $\mu^x(x_{<t}, z_{\le t})$ for more flexibility. 

Recall that the latent stochastic process $\{z_t\}$ (Eq. \eqref{eq:sv2}) in the SV model is by definition an i.i.d. Gaussian white noise process. We may extend this process to one with an inherent autoregressive dynamics and more flexibility that the mean $\mu^z(z_{<t})$ and variance $\Sigma^z(z_{<t})$ are functions of autoregressive structure on historical values. Hence, the generalised model can be formulated in the following framework:
\begin{align}
\label{eq:z}
z_t | z_{<t} &\sim \mathscr{N}(\mu^z(z_{<t}), \Sigma^z(z_{<t})),\\
\label{eq:x}
x_t | x_{<t}, z_{\le t} &\sim \mathscr{N}(\mu^x(x_{<t}, z_{\le t}), \Sigma^x(x_{<t}, z_{\le t})),
\end{align}
where we have presumed that both the observation $x_t$ and the latent variable $z_t$ are normally distributed. Note that the autoregressive process degenerates to i.i.d. white noise process when $\mu^z(z_{<t}) \equiv 0$ and $\Sigma^z(z_{<t}) \equiv \sigma^2_z$. It should be emphasised that the purpose of reinforcing an autoregressive structure \eqref{eq:z} of the latent variable $z_t$ is that we believe such formulation fits better to real scenarios from financial aspect compared with the i.i.d. convention: the price fluctuation of a certain stock is the consequence of not only its own history but also the influence from the environment, e.g. its competitors, up/downstream industries, relevant companies in the market, etc. Such external influence is ever-changing and may preserve memory and hence hard to characterise if restricted to i.i.d. noise. The latent variable $z_t$ with an autoregressive structure provides a possibility of decoupling the internal influential factors from the external ones, which we believe is the essence of introducing $z_t$.


\section{Neural Stochastic Volatility Models}

In this section, we establish the \emph{neural stochastic volatility model} (NSVM) for volatility estimation and prediction.


\subsection{Generating Observable Sequence}

Recall that the observable variable $x_t$ (Eq. \eqref{eq:x}) and the latent variable $z_t$ (Eq. \eqref{eq:z}) are described by autoregressive models (as $x_t$ also involves an exogenous input $z_{\le t}$). Let $p_{\ps{\Phi}}(x_t | x_{<t}, z_{\le t})$ and $p_{\ps{\Phi}}(z_t | z_{<t})$ denote the probability distributions of $x_t$ and $z_t$ at time $t$. The factorisation on the joint distributions of sequences $\{x_t\}$ and $\{z_t\}$ applies as follow:
{
\begin{align}
\label{eq:pz}
p_{\ps{\Phi}}(Z) &= \prod_t p_{\ps{\Phi}}(z_t | z_{<t})\notag \\
&= \prod_t \mathscr{N}(z_{t} ; \mu^z_{\ps{\Phi}}(z_{<t}), \Sigma^z_{\ps{\Phi}}(z_{<t})),\\
\label{eq:px|z}
p_{\ps{\Phi}}(X|Z) &= \prod_t p_{\ps{\Phi}}(x_t | x_{<t}, z_{\le t})\notag \\
&= \prod_t \mathscr{N}(x_{t} ; \mu^x_{\ps{\Phi}}(x_{<t}, z_{\le t}), \Sigma^x_{\ps{\Phi}}(x_{<t}, z_{\le t})),
\end{align}
}%
where $X = \{x_{1:T}\}$ and $Z = \{z_{1:T}\}$ represents the sequences of observable and latent variables, respectively, whereas $\ps{\Phi}$ stands for the collection of parameters of generative model. The unconditional generative model is defined as the joint distribution w.r.t. the latent variable $Z$ and observable $X$:
\begin{align}
\label{eq:pxz}
p_{\ps{\Phi}}(X, Z) =& \prod_t p_{\ps{\Phi}}(x_t | x_{<t}, z_{\le t}) p_{\ps{\Phi}}(z_t|z_{<t}).
\end{align}

It can be observed that the mean and variance are conditionally deterministic: given the historical information $\{z_{<t}\}$, the current mean $\mu^z_t = \mu^z_{\ps{\Phi}}(z_{<t})$ and variance $\Sigma^z_t = \Sigma^z_{\ps{\Phi}}(z_{<t})$ of $z_t$ is obtained and hence the distribution $\mathscr{N}(z_t; \mu^z_t, \Sigma^z_t)$ of $z_t$ is specified; after sampling $z_t$ from the specified distribution, we incorporate $\{x_{<t}\}$ and calculate the current mean $\mu^x_t = \mu^x_{\ps{\Phi}}(x_{<t}, z_{\le t})$ and variance $\Sigma^x_t = \Sigma^x_{\ps{\Phi}}(x_{<t}, z_{\le t})$ of $x_t$ and determine its distribution $\mathscr{N}(x_t; \mu^x_t, \Sigma^x_t)$ of $x_t$. It is natural and convenient to present such a procedure in a recurrent fashion because of its autoregressive nature. Since RNNs can essentially approximate arbitrary function of recurrent form, the means and variances, which may be driven by complex non-linear dynamics, can be efficiently computed using RNNs.

The unconditional generative model consists of two pairs of RNN and multi-layer perceptron (MLP), namely $\RNN^z_G$/$\MLP^z_G$ for the latent variable and $\RNN^x_G$/$\MLP^x_G$ for the observable. We stack those two RNN/MLP pairs together according to the causal dependency between variables. The unconditional generative model is implemented as the \emph{generative network} abstracted as follows:
\begin{align}
\label{eq:mlp_zg}
\{\mu^z_t, \Sigma^z_t\} &= \MLP^z_G(h^z_t; \ps{\Phi}),\\
\label{eq:rnn_zg}
h^z_t &= \RNN^z_G(h^z_{t-1}, z_{t-1}; \ps{\Phi}),\\
\label{eq:zg_t}
z_t &\sim \mathscr{N}(\mu^z_t, \Sigma^z_t),\\
\label{eq:mlp_xg}
\{\mu^x_t, \Sigma^x_t\} &= \MLP^x_G(h^x_t; \ps{\Phi}),\\
\label{eq:rnn_xg}
h^x_t &= \RNN^x_G(h^x_{t-1}, x_{t-1}, z_t; \ps{\Phi}),\\
\label{eq:xg_t}
x_t &\sim \mathscr{N}(\mu^x_t, \Sigma^x_t),
\end{align}
where $h^z_t$ and $h^x_t$ denote the hidden states of the corresponding RNNs. The MLPs map the hidden states of RNNs into the means and deviations of variables of interest. The collection of parameters $\ps{\Phi}$ is comprised of the weights of RNNs and MLPs. NSVM relaxes the conventional constraint that the latent variable $z_t$ is $\mathscr{N}(0,1)$ in a way that $z_t$ is no longer i.i.d noise but a time-varying signal from external process with self-evolving nature. As discussed above, this relaxation will benefit the effectiveness in real scenarios.

One should notice that when the latent variable $z_t$ is obtained, e.g. by inference (see details in the next subsection), the conditional distribution $p_{\ps{\Phi}}(X|Z)$ (Eq. \eqref{eq:px|z}) will be involved in generating the observable $x_t$ instead of the joint distribution $p_{\ps{\Phi}}(X, Z)$ (Eq. \eqref{eq:pxz}). This is essentially the scenario of predicting future values of the observable variable given its history. We will use the term ``generative model'' and will not discriminate the unconditional generative model or the conditional one as it can be inferred in context.


\subsection{Inferencing the Latent Process}

As the generative model involves the latent variable $z_t$, of which the true values are inaccessible even we have observed $x_t$, the marginal distribution $p_{\ps{\Phi}}(X)$ becomes the key that bridges the model and the data. However, the calculation of $p_{\ps{\Phi}}(X)$ itself or its complement, the posterior distribution $p_{\ps{\Phi}}(Z | X)$, is often intractable as complex integrals are involved. We are unable to learn the parameters by differentiating the marginal log-likelihood $\log p_{\ps{\Phi}}(X)$ or to infer the latent variables through the true posterior. Therefore, we consider instead a restricted family of tractable distributions $q_{\ps{\Psi}}(Z | X)$, referred to as the approximate posterior family, as approximations to the true posterior $p_{\ps{\Phi}}(Z | X)$ such that the family is sufficiently rich and of high capacity to provide good approximations.

It is straightforward to verify that given a sequence of observations $X=\{x_{1:T}\}$, for any $1\le t\le T$, $z_t$ is dependent on the entire observation sequences. Hence, we define the inference model with the spirit of mean-field approximation where the approximate posterior is Gaussian and the following factorisation applies:
{
\begin{align}
\label{eq:qz|x}
q_{\ps{\Psi}}(Z | X) &= \prod^T_{t=1} q_{\ps{\Psi}}(z_t | z_{<t}, x_{1:T})\notag \\
&= \prod_t \mathscr{N}(z_t ; \tilde{\mu}^z_{\ps{\Psi}}(z_{<t}, x_{1:T}), \tilde{\Sigma}^z_{\ps{\Psi}}(z_{<t}, x_{1:T})),
\end{align}
}%
where $\tilde{\mu}^z_{\ps{\Psi}}(z_{t-1}, x_{1:T})$ and $\tilde{\Sigma}^z_{\ps{\Psi}}(z_{t-1}, x_{1:T})$ are functions of the given observation sequence $\{x_{1:T}\}$, representing the approximated mean and variance of the latent variable $z_t$; $\ps{\Psi}$ denotes the collection of parameters of inference model.

The neural network implementation of the model, referred to as the \emph{inference network}, is designed to equip a cascaded architecture with an autoregressive RNN and a bidirectional RNN, where the bidirectional RNN incorporates both the forward and backward dependencies on the entire observations whereas the autoregressive RNN models the temporal dependencies on the latent variables:
\begin{align}
\label{eq:mlp_zi}
\{\tilde{\mu}^z_t, \tilde{\Sigma}^z_t\} &= \MLP^z_I(\tilde{h}^z_{t}; \ps{\Psi}),\\
\label{eq:rnn_zi}
\tilde{h}^z_{t} &= \RNN^z_I(\tilde{h}^z_{t-1}, z_{t-1}, [\tilde{h}^{\rightarrow}_t, \tilde{h}^{\leftarrow}_t]; \ps{\Psi}),\\
\label{eq:rnn_ri}
\tilde{h}^{\rightarrow}_t &= \RNN^{\rightarrow}_I(\tilde{h}^{\rightarrow}_{t-1}, x_t; \ps{\Psi}),\\
\label{eq:rnn_li}
\tilde{h}^{\leftarrow}_t &= \RNN^{\leftarrow}_I(\tilde{h}^{\leftarrow}_{t+1}, x_t; \ps{\Psi}),\\
\label{eq:zi_t}
z_t &\sim \mathscr{N}(\tilde{\mu}^z_t, \tilde{\Sigma}^z_t; \ps{\Psi}),
\end{align}
where $\tilde{h}^{\rightarrow}_t$ and $\tilde{h}^{\leftarrow}_t$ represent the hidden states of the forward and backward directions of the bidirectional RNN. The autoregressive RNN with hidden state $\tilde{h}^z_{t}$ takes the joint state $[\tilde{h}^{\rightarrow}_t, \tilde{h}^{\leftarrow}_t]$ of the bidirectional RNN and the previous value of $z_{t-1}$ as input. The inference mean $\tilde{\mu}^z_t$ and variance $\tilde{\Sigma}^z_t$ is computed by an MLP from the hidden state $\tilde{h}^z_t$ of the autoregressive RNN. We use the subscript $I$ instead of $G$ to distinguish the architecture used in inference model in contrast to that of the generative model. It should be emphasised that the inference network will collaborates with the generative network on conditional generating procedure.

\begin{algorithm}
\centering
\caption{Recursive Forecasting}\label{alg:recursiveforecast}
\begin{algorithmic}[1]
\Loop
\State $\{z^{\langle 1:S \rangle}_{1:t}\}$ $\gets$ draw $S$ paths from $q(z_{1:t} | x_{1:t})$
\State $\{z^{\langle 1:S \rangle}_{1:t+1}\}$ $\gets$ extend $\{z^{\langle 1:S \rangle}_{1:t}\}$ for 1 step via $p(z_{t+1} | z_{1:t})$
\State $\hat{p}(x_{t+1} | x_{1:t})$ $\gets$ $1/S\times \sum_{s} p(x_{t+1} | z^{\langle s \rangle}_{1:t+1}, x_{1:t})$
\State $\hat{\sigma}^2_{t+1} \gets \mathtt{var}\{\hat{x}^{1:S}_{t+1}\}$, $\{\hat{x}^{1:S}_{t+1}\} \sim \hat{p}(x_{\tau+1} | x_{1:\tau})$
\State $\{x_{1:t+1}\} \gets$ extend $\{x_{1:t}\}$ with new observation $x_{t+1}$
\State $t \gets t+1$, (optionally) retrain the model
\EndLoop
\end{algorithmic}
\end{algorithm}


\subsection{Forecasting Observations in Future}

For a volatility model to be practically applicable in forecasting, the generating procedure conditioning on the history is of essential interest. We start with 1-step-ahead prediction, which serves as building block of multi-step forecasting.

Given the historical observations $\{x_{1:T}\}$ up to time step $T$, 1-step-ahead prediction of either $\Sigma^x_{T+1}$ or $x_{T+1}$ is fully depicted by the conditional predictive distribution:
\begin{align}
\label{eq:1-step-ahead_exact}
p(x_{T+1} | x_{1:T}) &= \int_{z} p(x_{T+1} | z_{1:T+1}, x_{1:T})\notag \\
&\qquad \cdot p(z_{T+1} | z_{1:T}) p(z_{1:T} | x_{1:T})~\diff{z},
\end{align}
where the distributions on the right-hand side refer to those in the generative model with the generative parameters $\ps{\Phi}$ omitted. As the true posterior $p(z_{1:T} | x_{1:T})$ involved in Eq. \eqref{eq:1-step-ahead_exact} is intractable, the exact evaluation of conditional predictive distribution $p(x_{T+1} | x_{1:T})$ is difficult.

A straightforward solution is that we substitute the true posterior $p(z_{1:T} | x_{1:T})$ with the approximation $q(z_{1:T} | x_{1:T})$ (see Eq. \eqref{eq:qz|x}) and leverage $q(z_{1:T} | x_{1:T})$ to inference $S$ sample paths $\{z^{\langle 1:S \rangle}_{1:T}\}$ of the latent variables according to the historical observations $\{x_{1:T}\}$. The approximate posterior from a well-trained model is presumed to be a good approximation to the truth; hence the sample paths shall be mimics of the true but unobservable path. We then extend the sample paths one step further from $T$ to $T+1$ using the autoregressive generative distribution $p(z_{T+1} | z_{1:T})$ (see Eq. \eqref{eq:pz}). The conditional predictive distribution is thus approximated as
\begin{align}
\label{eq:1-step-ahead_approx}
\hat{p}(x_{T+1} | x_{1:T}) &\approx \frac{1}{S} \sum_{s} p(x_{T+1} | z^{\langle s \rangle}_{1:T+1}, x_{1:T}),
\end{align}
which is essentially a mixture of $S$ Gaussians. In the case of multi-step forecasting, a common solution in practice is to perform a recursive 1-step-ahead forecasting routine with model updated as new observation comes in; the very same procedure can be applied except that more sample paths should be evaluated due to the accumulation of uncertainty. Algorithm~\ref{alg:recursiveforecast} gives the detailed rolling scheme.


\section{Experiment}

In this section, we present the experiment on real-world stock price time series to validate the effectiveness and to evaluate the performance of the prosed model.


\subsection{Dataset and Pre-processing}

The raw dataset comprises 162 univariate time series of the daily closing stock price, chosen from China's A-shares and collected from 3 institutions. The choice is made by selecting those with earlier listing date of trading (from 2006 or earlier) and fewer suspension days (at most 50 suspension days within the entire period of observation), such that the undesired noises introduced by insufficient observation or missing values -- highly influential on the performance but essentially irrelevant to the purpose of volatility modelling -- can be reduced to the minimum. The raw price series is cleaned by aligning and removing abnormalities: we manually aligned the mismatched part and interpolated the missing value by stochastic regression imputation \cite{little2014statistical} where the imputed value is drawn from a Gaussian distribution with mean and variance calculated by regression on the empirical value within a short interval of 20 recent days. The series is then transformed from actual prices $s_t$ into log-returns $x_t = \log(s_t/s_{t-1})$ and normalised. Moreover, we combinatorically choose a predefined number $d$ out of 162 univariate log-return series and aggregate the selected series at each time step to form a $d$-dimensional multivariate time series, the choice of $d$ is in accordance with the rank of correlation, e.g. $d=6$ in our experiments. Theoretically, it leads to a much larger volume of data as ${{162}\choose{6}} > 2\times 10^{10}$. Specifically, the actual dataset for training and evaluation comprises a collection of 2000 series of $d$-dimensional normalised log-return vectors of length $2570$ ($\sim$ 7 years) with no missing values. We divide the whole dataset into two subsets for training and testing along the time axis: the first 2000 time steps of each series have been used as training samples whereas the rest 570 steps of each series as the test samples. 


\subsection{Baselines}

We select several deterministic volatility models from the GARCH family as baselines:
{
\begin{enumerate}
\setlength\itemsep{0em}
\item{Quadratic models}
\begin{itemize}
\setlength\itemsep{0em}
\item{ARCH(1); GARCH(1,1); GJR-GARCH(1,1,1);}
\end{itemize}
\setlength\itemsep{0em}
\item{Absolute value models}
\begin{itemize}
\setlength\itemsep{0em}
\item{AVARCH(1); AVGARCH(1,1); TARCH(1,1,1);}
\end{itemize}
\setlength\itemsep{0em}
\item{Exponential models.}
\begin{itemize}
\setlength\itemsep{0em}
\item{EARCH(1); EGARCH(1,1);}
\end{itemize}
\end{enumerate}
}%
\noindent Moreover, two stochastic volatility models are compared: 
\begin{enumerate}
\setlength\itemsep{0em}
\item{MCMC volatility model: \emph{stochvol};}
\item{Gaussian process volatility model \emph{GP-Vol}.}
\end{enumerate}
For the listed models, we retrieve the authors' implementations or tools: \emph{stochvol}\footnote{\scriptsize\url{https://cran.r-project.org/web/packages/stochvol}}, \emph{GP-Vol}\footnote{\scriptsize\url{http://jmhl.org}} (the hyperparameters are chosen as suggested in \cite{DBLP:conf/nips/WuHG14}) and implement the models, such as GARCH, EGARCH, GJR-GARCH, etc., based on several widely-used packages\footnote{\scriptsize\url{https://pypi.python.org/pypi/arch/4.0}}\footnote{\scriptsize\url{https://www.kevinsheppard.com/MFE_Toolbox}}\footnote{\scriptsize\url{https://cran.r-project.org/web/packages/fGarch}} for time series analysis. All baselines are evaluated in terms of the negative log-likelihood on the test samples, where 1-step-ahead forecasting is carried out in a recursive fashion similar to Algorithm \ref{alg:recursiveforecast}. 

\begin{table*}[t]
\setlength{\tabcolsep}{3pt}
\centering
\caption{The performance of the proposed model and the baselines in terms of negative log-likelihood (NLL) evaluated on the test samples of real-world stock price time series: each row from 1 to 10 lists the average NLL for a specific individual stock; the last row summarises the average NLL of the entire test samples of all 162 stocks.
}
\label{tbl:performance}
{\footnotesize
\begin{tabular}{rrrrrrrrrrrrr}
\Xhline{1.2pt}
Stock & NSVM-corr & NSVM-diag & ARCH & GARCH & GJR & AVARCH & AVGCH & TARCH & EARCH & EGARCH & stochvol & GP-Vol\Tstrut\Bstrut\\
\Xhline{0.7pt}
1 & \bf{1.11341} & 1.42816 & 1.36733 & 1.60087 & 1.60262 & 1.34792 & 1.57115 & 1.58156 & 1.33528 & 1.53651 & 1.39638 & 1.56260\Tstrut\\
2 & \bf{1.04058} & 1.28639 & 1.35682 & 1.63586 & 1.59978 & 1.32049 & 1.46016 & 1.45951 & 1.35758 & 1.52856 & 1.37080 & 1.47025 \\
3 & \bf{1.03159} & 1.32285 & 1.37576 & 1.44640 & 1.45826 & 1.34921 & 1.44437 & 1.45838 & 1.33821 & 1.41331 & 1.25928 & 1.48203 \\
4 & \bf{1.06467} & 1.32964 & 1.38872 & 1.45215 & 1.43133 & 1.37418 & 1.44565 & 1.44371 & 1.35542 & 1.40754 & 1.36199 & 1.32451 \\
5 & \bf{0.96804} & 1.22451 & 1.39470 & 1.31141 & 1.30394 & 1.37545 & 1.28204 & 1.27847 & 1.37697 & 1.28191 & 1.16348 & 1.41417 \\
6 & \bf{0.96835} & 1.23537 & 1.44126 & 1.55520 & 1.57794 & 1.39190 & 1.47442 & 1.47438 & 1.36163 & 1.48209 & 1.15107 & 1.24458 \\
7 & \bf{1.13580} & 1.43244 & 1.36829 & 1.65549 & 1.71652 & 1.32314 & 1.50407 & 1.50899 & 1.29369 & 1.64631 & 1.42043 & 1.19983 \\
8 & \bf{1.03752} & 1.26901 & 1.39010 & 1.47522 & 1.51466 & 1.35704 & 1.44956 & 1.45029 & 1.34560 & 1.42528 & 1.26289 & 1.47421 \\
9 & \bf{0.95157} & 1.15896 & 1.42636 & 1.32367 & 1.24404 & 1.42047 & 1.35427 & 1.34465 & 1.42143 & 1.32895 & 1.12615 & 1.35478 \\
10 & \bf{0.99105} & 1.13143 & 1.36919 & 1.55220 & 1.29989 & 1.24032 & 1.06932 & 1.04675 & 23.35983 & 1.20704 & 1.32947 & 1.18123\Bstrut\\
\Xhline{0.7pt}
AVG & \bf{1.18354} & 1.23521 & 1.27062 & 1.27051 & 1.28809 & 1.28827 & 1.27754 & 1.29010 & 1.33450 & 1.36465 & 1.27098 & 1.34751\Tstrut\Bstrut\\
\Xhline{1.2pt}
\end{tabular}
}%
\end{table*}


\subsection{Model Implementation}

In our experiments, we predefine the dimensions of observable variables to be $\dim{x_t} = 6$ and the latent variables $\dim{z_t} = 4$. Note that the dimension of the latent variable is smaller than that of the observable, which allows us to extract a compact representation. The NSVM implementation in our experiments is composed of two neural networks, namely the generative network (see Eq. \eqref{eq:mlp_zg}-\eqref{eq:xg_t}) and inference network (see Eq. \eqref{eq:mlp_zi}-\eqref{eq:zi_t}). Each RNN module contains one hidden layer of size $10$ with GRU cells; MLP modules are 2-layered fully-connected feedforward networks, where the hidden layer is also of size $10$ whereas the output layer splits into two equal-sized sublayers with different activation functions: one applies exponential function to ensure the non-negativity for variance while the other uses linear function to calculate mean estimates. Thus $\MLP^z_I$'s output layer is of size $4+4$ for $\{\tilde{\mu}^z,\tilde{\Sigma}^z\}$ whereas the size of $\MLP^x_G$'s output layer is $6+6$ for $\{\mu^x,\Sigma^x\}$. During the training phase, the inference network is connected with the conditional generative network (see, Eq. \eqref{eq:mlp_zg}-\eqref{eq:zg_t}) to establish a bottleneck structure, the latent variable $z_t$ inferred by variational inference \cite{DBLP:journals/corr/KingmaW13,DBLP:conf/icml/RezendeMW14} follows a Gaussian approximate posterior; the size of sample paths is set to $S=100$. The parameters of both networks are jointly learned, including those for the prior. We introduce Dropout \cite{DBLP:journals/jmlr/SrivastavaHKSS14} into each RNN modules and impose $L2$-norm on the weights of MLP modules as regularistion to prevent overshooting; Adam optimiser \cite{DBLP:journals/corr/KingmaB14} is exploited for fast convergence; exponential learning rate decay is adopted to anneal the variations of convergence as time goes. Two covariance configurations are adopted: 1. we stick with diagonal covariance matrices configurations; 2. we start with diagonal covariance and then apply rank-1 perturbation \cite{DBLP:conf/icml/RezendeMW14} during fine-tuning until training is finished. The recursive 1-step-ahead forecasting routine illustrated as Algorithm \ref{alg:recursiveforecast} is applied in the experiment for both training and test phase: during the training phase, a single NSVM is trained, at each time step, on the entire training samples to learn a holistic dynamics, where the latent shall reflect the evolution of environment; in the test phase, on the other hand, the model is optionally retrained, at every 20 time steps, on each particular input series of the test samples to keep track on the specific trend of that series. In other words, the trained NSVM predicts 20 consecutive steps before it is retrained using all historical time steps of the input series at present. Correspondingly, all baselines are trained and tested at every time step of each univariate series using standard calibration procedures. The negative log-likelihood on test samples has been collected for performance assessment. We train the model on a single-GPU (Titan X Pascal) server for roughly two hours before it converges to a certain degree of accuracy on the training samples. Empirically, the training phase can be processed on CPU in reasonable time, as the complexity of the model as well as the size of parameters is moderate.

\vspace{-.5em}
\begin{figure}[!t]
\centering
\subfloat[The volatility forecasting for Stock 37.]{\includegraphics[width=0.90\columnwidth]{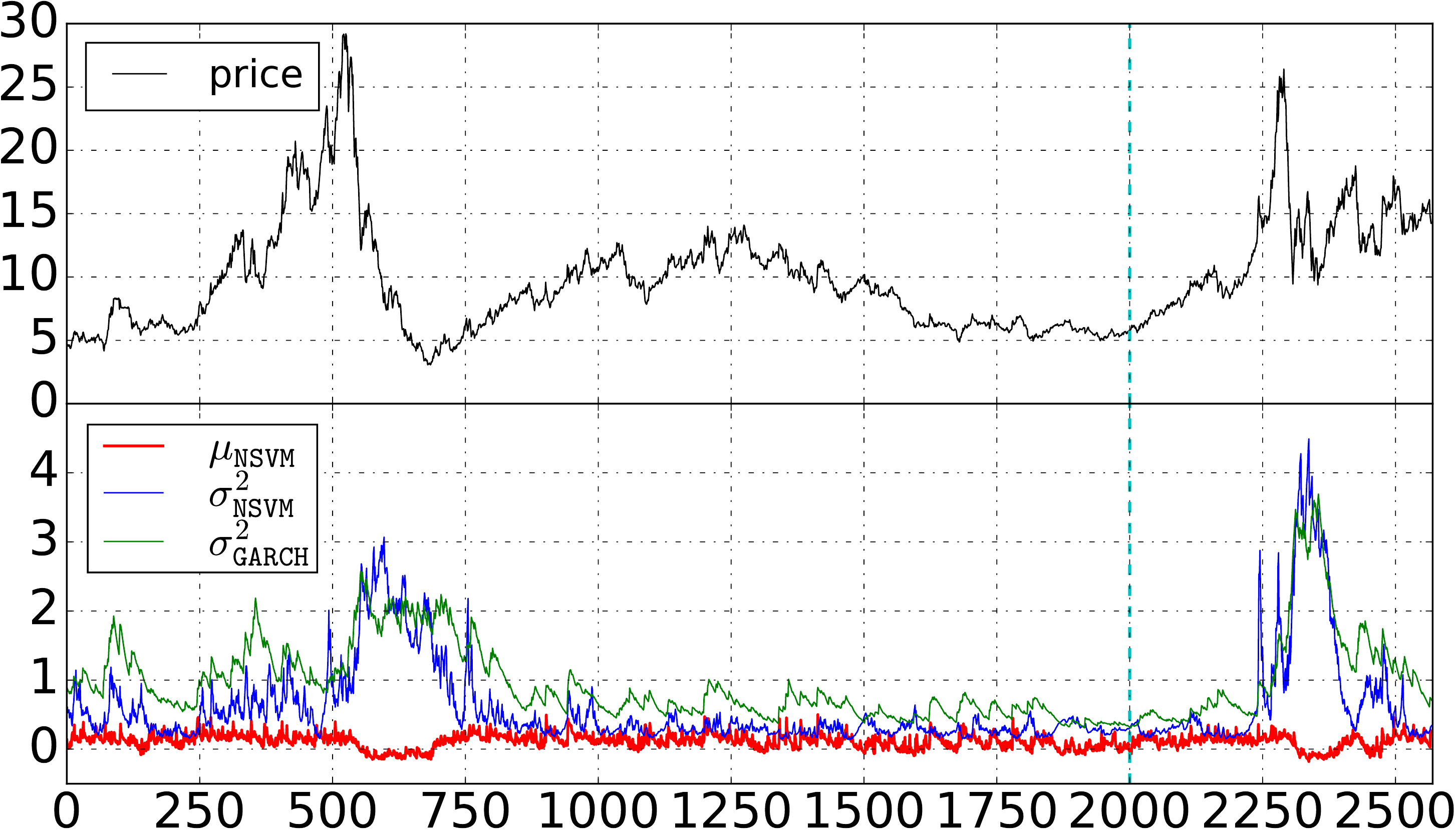}
\label{fig:case1}}
\hfil
\subfloat[The volatility forecasting for Stock 82.]{\includegraphics[width=0.90\columnwidth]{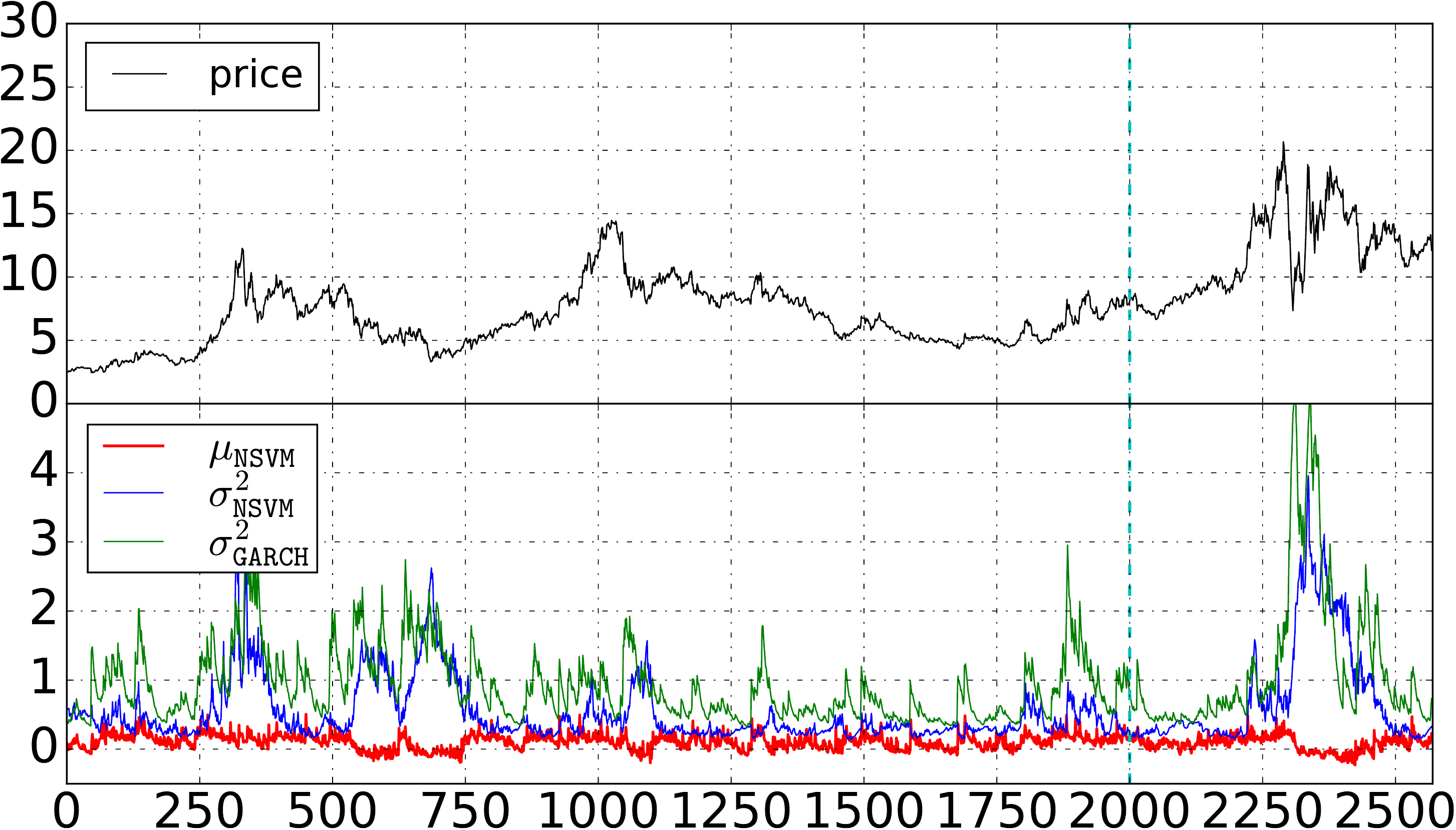}
\label{fig:case2}}
\caption{Case studies of volatility forecasting.}
\label{fig:casestudy}
\end{figure}
\vspace{-.5em}

\subsection{Result and Discussion}

The performance of NSVM and baselines is listed for comparison in Table \ref{tbl:performance}: the performance on the first 10 individual stocks (chosen in alphacetical order but anonymised here) and the average score on all 162 stocks are reported in terms of negative log-likelihood (NLL) measure.
The result shows that NSVM has achieved higher accuracy over the baselines on the task of volatility modelling and forecasting on NLL, which validates the high flexibility and rich expressive power of NSVM for volatility modelling and forecasting. In particular, NSVM with rank-1 perturbation (referred to as NSVM-corr in Table \ref{tbl:performance}) beats all other models in terms of NLL, while NSVM with diagonal covariance matrix (i.e. NSVM-diag) outperforms GARCH(1,1) on 142 out of 162 stocks. Although the improvement comes at the cost of longer training time before convergence, it can be mitigated by applying parallel computing techniques as well as more advanced network architecture or training methods. Apart from the higher accuracy NSVM obtained, it provides us with a rather general framework to generalise univariate time series models of any specific functional form to the corresponding multivariate cases by extending network dimensions and manipulating the covariance matrices. A case study on real-world financial datasets is illustrated in Fig.~\ref{fig:casestudy}. 

NSVM shows higher sensibility on drastic changes and better stability on moderate fluctuations: the response of NSVM in Fig.~\ref{fig:case1} is more stable in $t\in [1600, 2250]$, the period of moderate price fluctuation; while for drastic price change at $t=2250$, the model responds with a sharper spike compared with the quadratic GARCH model. Furthermore, NSVM demonstrates the inherent non-linearity in both Fig.~\ref{fig:case1} and \ref{fig:case2}: at each time step within $t\in [1000, 2000]$, the model quickly adapts to the current fluctuation level whereas GARCH suffers from a relatively slower decay from the previous influences.
The cyan vertical line at $t=2000$ splits the training samples and test samples. We show only one instance within our dataset due to the limitation of pages, the performance of other instances are similar.

\section{Conclusion}

In this paper, we proposed a new volatility model, referred to as NSVM, for volatility estimation and prediction. We integrated statistical models with deep neural networks, leveraged the characteristics of each model, organised the dependences between random variables in the form of graphical models, implemented the mappings among variables and parameters through RNNs and MLPs, and finally established a powerful stochastic recurrent model with universal approximation capability. The proposed architecture comprises a pair of complementary stochastic neural networks: the generative network and inference network. The former models the joint distribution of the stochastic volatility process with both observable and latent variables of interest; the latter provides with the approximate posterior i.e. an analytical approximation to the (intractable) conditional distribution of the latent variables given the observable ones. The parameters (and consequently the underlying distributions) are learned (and inferred) via variational inference, which maximises the lower bound for the marginal log-likelihood of the observable variables. NSVM has presented higher accuracy on the task of volatility modelling and forecasting on real-world financial datasets, compared with various widely-used models, such as GARCH, EGARCH, GJR-GARCH, TARCH in the GARCH family, MCMC volatility model \emph{stochvol} as well as Gaussian process volatility model \emph{GP-Vol}. Future work on NSVM would be to investigate the modelling of time series with non-Gaussian residual distributions, in particular the heavy-tailed distributions e.g. LogNormal $\log\mathscr{N}$ and Student's $t$-distribution.


\clearpage
{\small
\bibliographystyle{aaai}
\bibliography{nsvm}
}%

\clearpage
\appendix

\section{Learning Parameters / Calibration}

Given the observations $X=\{x_{1:T}\}$, we target at maximising the marginal log-likelihood $p_{\ps{\Phi}}(X)$ w.r.t. $\ps{\Phi}$, where the actual posterior $p_{\ps{\Phi}}(Z|X)$ is involved. Because of the intractability of $p_{\ps{\Phi}}(Z|X)$, the exact inference is not applicable; we have to seek for approximate solutions instead. 

We factorise the marginal log-likelihood $\log{p_{\ps{\Phi}}(X)}$ as
\begin{align}
\log{p_{\ps{\Phi}}(X)} &= \mathbb{E}_{q_{\ps{\Psi}}(Z|X)} \bigg[ \log{\frac{p_{\ps{\Phi}}(X,Z)}{p_{\ps{\Phi}}(Z|X)}} \bigg]\notag \\
&= \mathbb{E}_{q_{\ps{\Psi}}(Z|X)} \bigg[ \log{\bigg( \frac{p_{\ps{\Phi}}(X,Z)}{q_{\ps{\Psi}}(Z|X)} \cdot \frac{q_{\ps{\Psi}}(Z|X)}{p_{\ps{\Phi}}(Z|X)} \bigg)} \bigg]\notag \\
&= \mathcal{L}[q; X, \ps{\Phi}, \ps{\Psi}] + KL[ q_{\ps{\Psi}}(Z | X) \| p_{\ps{\Phi}}(Z | X) ]\notag \\
&\ge \mathcal{L}[q; X, \ps{\Phi}, \ps{\Psi}],\notag \\
\label{eq:elbo}
\mbox{where}~~\mathcal{L}[&q; X, \ps{\Phi}, \ps{\Psi}] = \mathbb{E}_{q_{\ps{\Psi}}(Z | X)} \bigg[ \log{\frac{ p_{\ps{\Phi}}(X, Z)}{q_{\ps{\Psi}}(Z | X)}} \bigg].
\end{align}
Note that we have introduced a tractable, $\ps{\Psi}$-parameterised distribution $q_{\ps{\Psi}}(Z|X)$ from a flexible family of distributions to approximate the actual posterior $p_{\ps{\Phi}}(Z|X)$. The evidence lower bound (ELBO) $\mathcal{L}[q; X, \ps{\Phi}, \ps{\Psi}]$ in Eq. \eqref{eq:elbo} is essentially a functional w.r.t. $q$, conditioning on the observations $X$ and parameterised by the parameter sets $\ps{\Phi}, \ps{\Psi}$ of both generative and inference models. Theoretically, ELBO ensures a lower bound on the marginal log-likelihood, and can be maximised via gradient-based optimisers.

It is usually the case that Eq. \eqref{eq:elbo} lacks a closed-form expression. We have to estimate the ELBO using Monte Carlo integration on the latent variable $z_t$. Provided $S$ sample paths drawn by the inference model defined in Eq. \eqref{eq:qz|x}, the estimator of ELBO can be calculated as the path average:
{\small
\vskip -1.0em
\begin{align*}
\widehat{\mathcal{L}}\big[q; \{z^{\langle 1:S \rangle}_{1:T}\}, X, \ps{\Phi}, \ps{\Psi}\big] = \frac{1}{S} \sum_{s,t} \Bigg[ \log\frac{p_{\ps{\Phi}}(x_t | x_{<t}, z^{\langle s \rangle}_{\le t}) p_{\ps{\Phi}}(z^{\langle s \rangle}_t|z^{\langle s \rangle}_{<t})}{q_{\ps{\Psi}}(z^{\langle s \rangle}_t | z^{\langle s \rangle}_{<t}, x_{1:T})} \Bigg],
\end{align*}
\vskip -.5em\noindent
}%
where $\{z^{\langle 1:S \rangle}_{1:T}\}$ denotes the collection of $S$ sample paths. 

\begin{figure}[!b]
\vskip -1.0em
\centering
\includegraphics[width=0.48\linewidth]{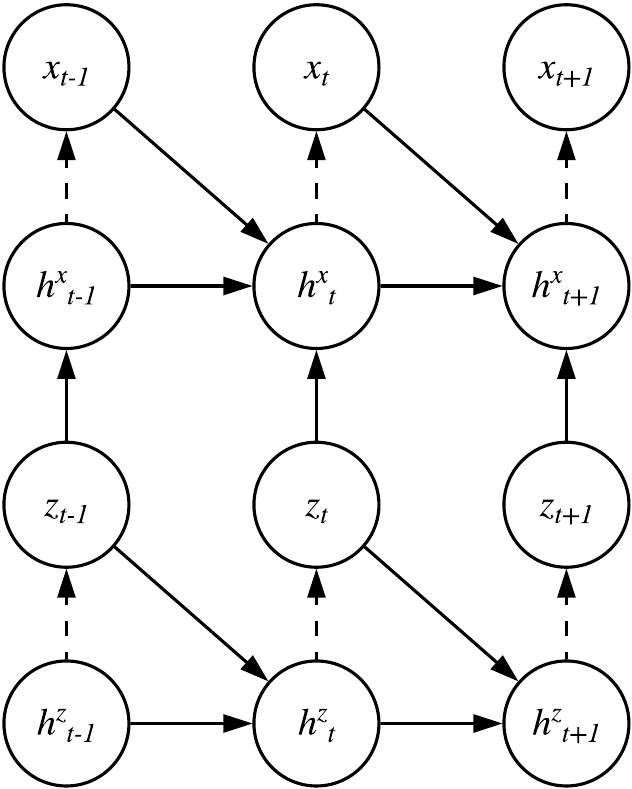}
\caption{Generalised stochastic volatility model.}
\label{fig:gennet}
\end{figure}

By assuming the latent variable $z_t$ being Gaussian, we can readily apply the reparameterisation technique \cite{DBLP:journals/corr/KingmaW13} to $z_t$ to form an unbiased gradient estimator:
{
\begin{align}
\label{eq:deriv_est}
\nabla_{\ps{\Phi}, \ps{\Psi}} \widehat{\mathcal{L}}\big[\{z^{\langle s \rangle}_t = \tilde{\mu}^{z}_t + \tilde{A}^{z}_t \epsilon^{\langle s \rangle}_t\}^{\langle 1:S \rangle}_{1:T}\big],
\end{align}
}%
where $\epsilon^{\langle s \rangle}_t \sim \mathscr{N}(0, I_z)$ is the standard Gaussian variable.

The reparameterisation extracts the randomness out of the latent variable $z^{\langle s \rangle}_t$ via $\epsilon^{\langle s \rangle}_t$, leaving $\tilde{\mu}^{z}_t$ and $\tilde{A}^{z}_t$, i.e. the mean and standard deviation of $z^{\langle s \rangle}_t$, being deterministic functions. It guarantees that the gradient-based optimisation techniques are applicable by isolating the model parameters ($\tilde{\mu}^{z}_t$ and $\tilde{A}^{z}_t$) from the sampling procedure (involving $\epsilon^{\langle s \rangle}_t$). 

\begin{figure}[!h]
\centering
\subfloat[Autoregressive network for latent variable $z_t$.]{\includegraphics[width=0.55\columnwidth]{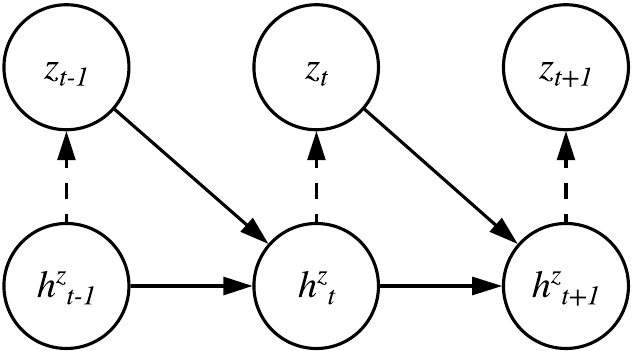}
\label{fig:gennet-component1}}
\hfil
\subfloat[Conditional generative network for observable variable $x_t$.]{\includegraphics[width=0.55\columnwidth]{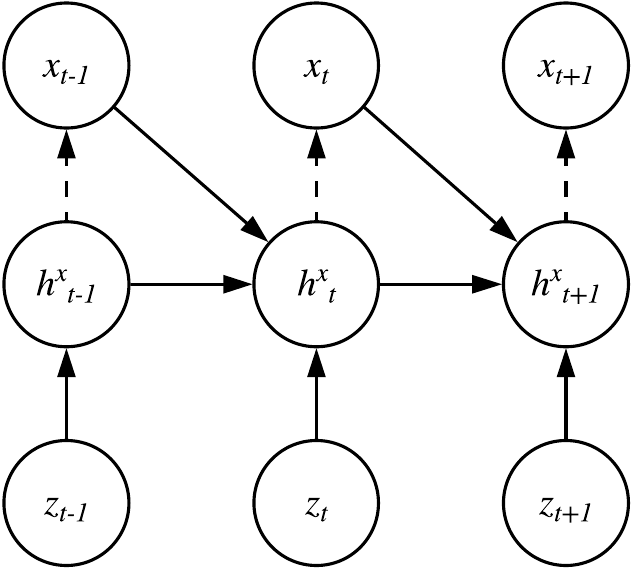}
\label{fig:gennet-component2}}
\vskip -0.5em
\caption{Two components of the generative network.}
\label{fig:gennet-components}
\vskip -1.5em
\end{figure}

\section{Illustrations of stochastic volatility modelling, training and forecasting}

Recall Eq. \eqref{eq:z} and \eqref{eq:x}, the generalised formulation for stochastic volatility modelling reads
\vskip -1.35em
\begin{align*}
z_t | z_{<t} &\sim \mathscr{N}(\mu^z(z_{<t}), \Sigma^z(z_{<t})),\\
x_t | x_{<t}, z_{\le t} &\sim \mathscr{N}(\mu^x(x_{<t}, z_{\le t}), \Sigma^x(x_{<t}, z_{\le t})).
\end{align*}
\vskip -0.35em

By introducing hidden state $h^z_t$ and $h^x_t$ as memory for historical information integration, the formulation is essentially equivalent to the recurrent model illustrated as Fig. \ref{fig:gennet}.

We decompose the recurrent model into two components in a similar way as one would apply in factorising $p_{\ps{\Phi}}(X, Z)$ into the marginal distribution $p_{\ps{\Phi}}(Z)$ in Eq. \eqref{eq:pz} and the conditional distribution $p_{\ps{\Phi}}(X|Z)$ in Eq. \eqref{eq:px|z}.

The marginal $p_{\ps{\Phi}}(Z)$ is implemented by
\begin{align*}
\{\mu^z_t, \Sigma^z_t\} &= \MLP^z_G(h^z_t; \ps{\Phi}),\\
h^z_t &= \RNN^z_G(h^z_{t-1}, z_{t-1}; \ps{\Phi}),\\
z_t &\sim \mathscr{N}(\mu^z_t, \Sigma^z_t),
\end{align*}
which represents an autoregressive network for the latent $z_t$ as illustrated in Fig. \ref{fig:gennet-component1}. The conditional $p_{\ps{\Phi}}(X|Z)$ is built as
\begin{align*}
\{\mu^x_t, \Sigma^x_t\} &= \MLP^x_G(h^x_t; \ps{\Phi}),\\
h^x_t &= \RNN^x_G(h^x_{t-1}, x_{t-1}, z_t; \ps{\Phi}),\\
x_t &\sim \mathscr{N}(\mu^x_t, \Sigma^x_t),
\end{align*}
which corresponds to a conditional generative network for the observable $x_t$ as in Fig. \ref{fig:gennet-component2}. 

\begin{figure*}[!t]
\centering
\includegraphics[width=0.725\linewidth]{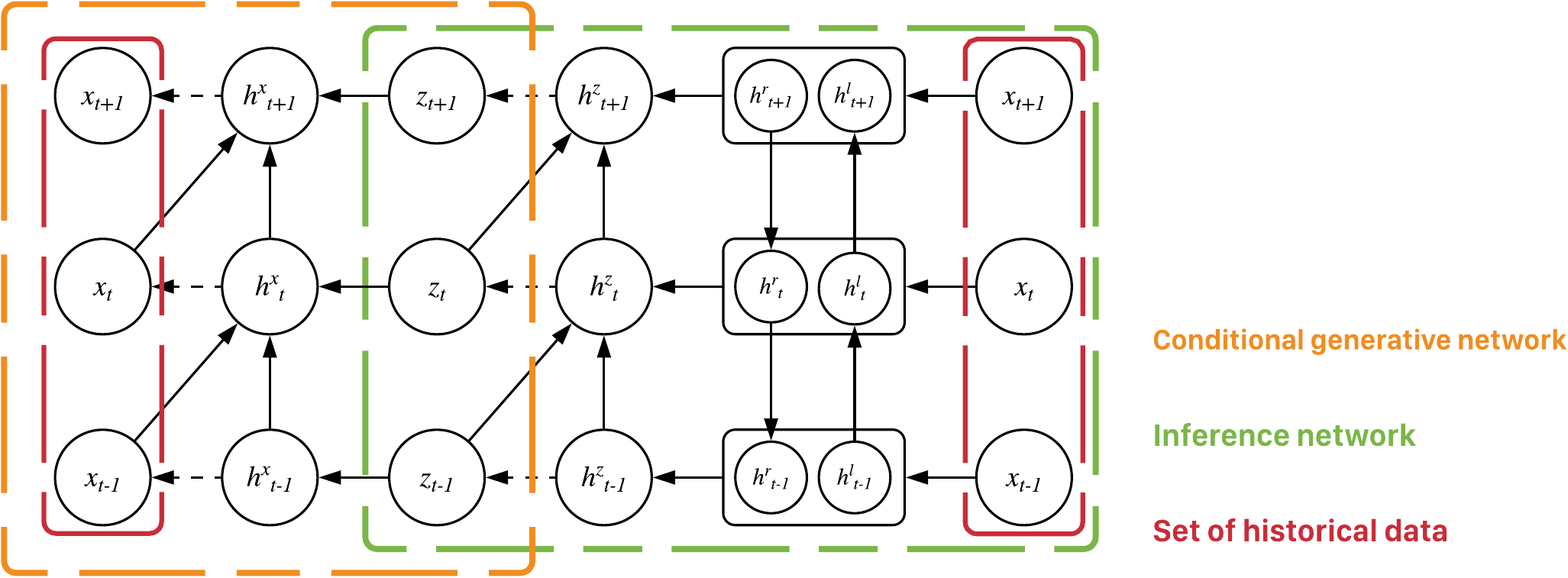}
\caption{Illustration of the training setup.}
\label{fig:training}
\end{figure*}

\begin{figure}[!h]
\centering
\includegraphics[width=0.55\linewidth]{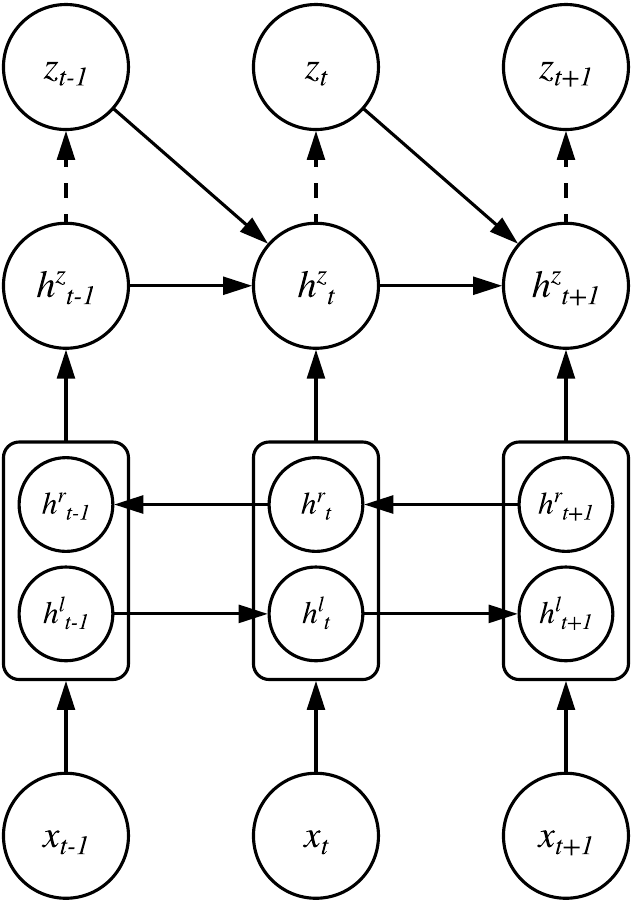}
\caption{Architecture of the inference network.}
\label{fig:infnet}
\vskip -1em
\end{figure}

On the other hand, the inference network is implemented in a similar recurrent fashion, as an autoregressive network with bidirectional dependencies: 
\begin{align*}
\{\tilde{\mu}^z_t, \tilde{\Sigma}^z_t\} &= \MLP^z_I(\tilde{h}^z_{t}; \ps{\Psi}),\\
\tilde{h}^z_{t} &= \RNN^z_I(\tilde{h}^z_{t-1}, z_{t-1}, [\tilde{h}^{\rightarrow}_t, \tilde{h}^{\leftarrow}_t]; \ps{\Psi}),\\
\tilde{h}^{\rightarrow}_t &= \RNN^{\rightarrow}_I(\tilde{h}^{\rightarrow}_{t-1}, x_t; \ps{\Psi}),\\
\tilde{h}^{\leftarrow}_t &= \RNN^{\leftarrow}_I(\tilde{h}^{\leftarrow}_{t+1}, x_t; \ps{\Psi}).\\
z_t &\sim \mathscr{N}(\tilde{\mu}^z_t, \tilde{\Sigma}^z_t; \ps{\Psi}),
\end{align*}
The architecture of inference network is illustrated in Fig. \ref{fig:infnet}.

The training procedure involves the inference network (in Fig. \ref{fig:infnet}) and the conditional generative network (in Fig. \ref{fig:gennet-component2}); the autoregressive network (in Fig. \ref{fig:gennet-component1}) will not be utilised.

The historical observations $\{x_{<t}\}$ is fed into the inference network, which outputs the inferred sequence $\{z_{<t}\}$ of the causing latent variable. The latent sequence $\{z_{<t}\}$ is then put into the conditional generative network (in Fig. \ref{fig:gennet-component2}) to generate the predictions $\{\hat{x}_{<t}\}$ . The likelihood of the predictions $\{\hat{x}_{<t}\}$ regarding the actual observations $\{x_{<t}\}$ is calculated and the networks are optimised via gradient-based methods. Figure \ref{fig:training} illustrates the setting of networks during training. 

\begin{figure}[!b]
\centering
\includegraphics[width=0.95\linewidth]{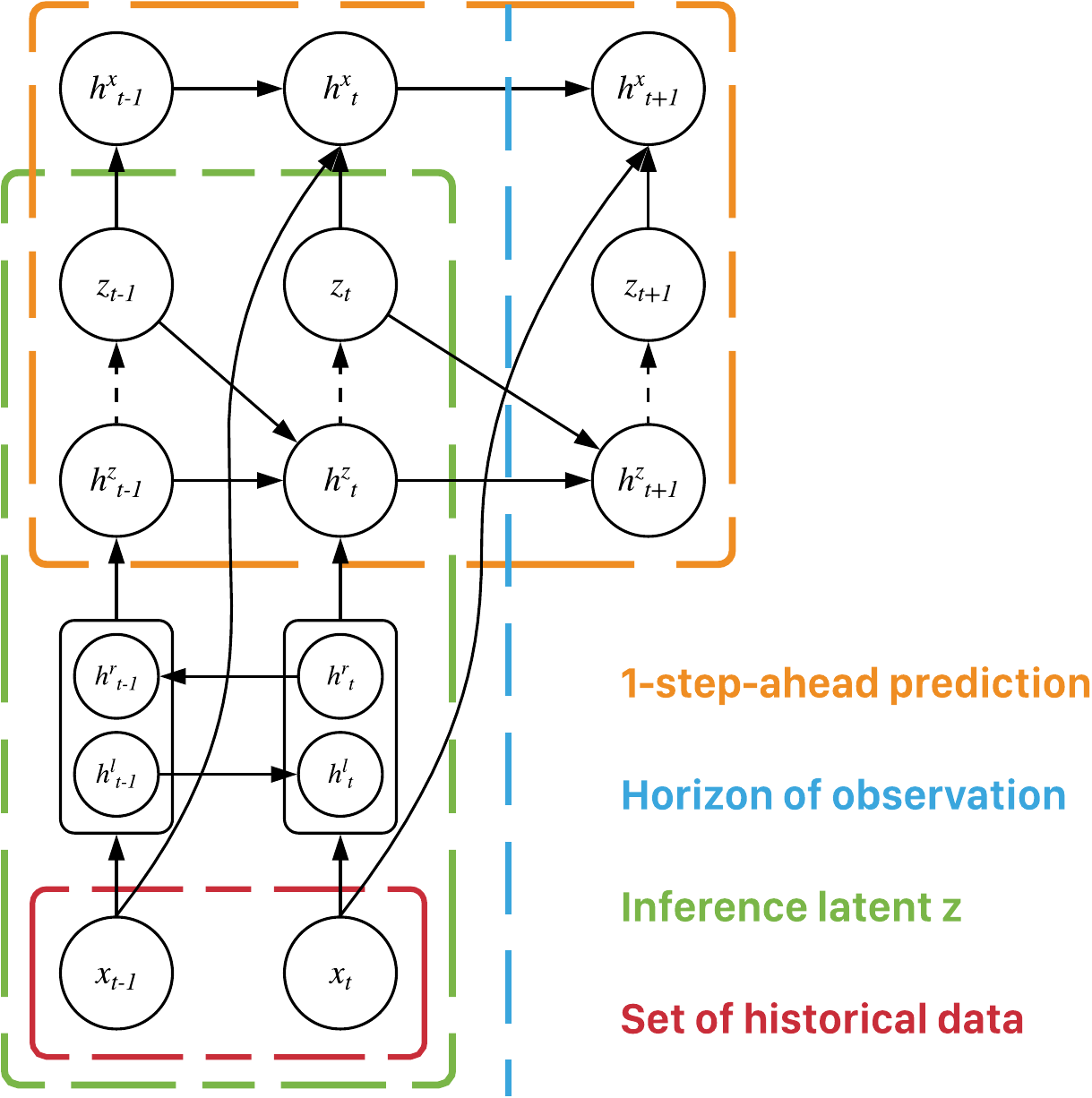}
\caption{Illustration of the forecasting setup.}
\label{fig:forecasting}
\end{figure}

The procedure of forecasting contains three steps: 1. feeding the historical data as the observations $\{x_{<t}\}$ into the inference network (in Fig. \ref{fig:infnet}) to infer the latent variables $\{z_{<t}\}$ that might have caused that observations, 2. evolving the latent dynamics using the autoregressive network (in Fig. \ref{fig:gennet-component2}) with the inferred sequence $\{z_{<t}\}$ to produce the latent variable $z_t$ for the next time step, and 3. invoking the conditional generative network (in Fig. \ref{fig:gennet-component2}) with $\{z_{1:t}\}$ and $\{x_{<t}\}$ to predict the next possible $x_t$. The procedure is shown in Fig. \ref{fig:forecasting}.

\begin{algorithm}
\centering
\caption{Computation scheme for rank-$K$ perturbation}\label{alg:perturbation}
\begin{algorithmic}[1]
\Require{full-rank diagonal $D$; rank-$K$ matrix $V = [v_k]_1^K$}
\Ensure{factor matrix $A$ s.t. $AA^\top = (D + VV^\top)^{-1} = \Sigma$}
\State $A \gets D^{-\frac{1}{2}}$
\For{$k\gets 1, K$}
\State $\gamma \gets v_k^\top AA^\top v_k$
\State $\eta \gets (1+\gamma)^{-1}$
\State $A \gets A - [(1-\sqrt{\eta})/\gamma] AA^\top v_k v_k^\top A$
\EndFor
\end{algorithmic}
\end{algorithm}

\section{Covariance Parameterisation}

So far we have kept the covariance matrix $\Sigma$ in our formulas to allow for multivariate forecasting. It entails a complexity of computation of order $\mathcal{O}(M^3)$ to maintain and update the full-size covariance matrix $\Sigma$ of $M$ dimensions; for cases in high dimensions, the computational cost for the full-size $\Sigma$ becomes unaffordable. Thus, we have to seek for alternatives that are more economic in terms of computation. A practical approach is to leverage low-rank perturbations on diagonal matrices $D$ such that $\Sigma^{-1} = D + VV^\top$, where $V = [v_{1:K}]$ is the rank-$K$ perturbation with each $v_k$ being independent $M$-dimensional column vector. The corresponding covariance matrix and its determinant can be readily calculated using \emph{Woodbury identity} and \emph{matrix determinant lemma}:
\begin{align}
\Sigma = D^{-1} &- D^{-1}V(I+V^\top D^{-1}V)^{-1}V^\top D^{-1}, \\
\ln\det{\Sigma} &= -\ln\det{(D + VV^\top)}\notag \\
&= -\ln\det{D} - \ln\det{(I+V^\top D^{-1}V)}.
\end{align}

To solve the standard deviation $A$ in the factorisation $\Sigma = AA^\top$, we start with rank-$1$ perturbation. Given $K=1$, matrix $V=[v]$ is basically a column vector, hence $I+V^\top D^{-1}V = 1+v^\top D^{-1}v$ returns a real number. A solution of $A$ reads
\begin{align*}
A = D^{-\frac{1}{2}} - [(1-\sqrt{\eta})/\gamma]D^{-1}vv^\top D^{-\frac{1}{2}},
\end{align*}
where $\gamma = v^\top D^{-1}v$ and $\eta = (1+\gamma)^{-1}$. The complexity of computation is just of order $\mathcal{O}(M)$. Observe that $VV^\top = \sum_{k=1}^K v_k v_k^\top$, the perturbation of rank $K$ is merely the superposition of $K$ rank-$1$ perturbations. Thus, we can calculate $A$ in a recurrent fashion, where the complexity of computation for rank-$K$ perturbation remains $\mathcal{O}(M)$ when $K\ll M$ holds. Algorithm~\ref{alg:perturbation} describes the detailed scheme of computation.

\end{document}